\documentclass{article}

\PassOptionsToPackage{numbers, compress}{natbib}

\usepackage{graphicx}

\usepackage[final]{tackling_climate_workshop_style}

\usepackage[utf8]{inputenc} 
\usepackage[T1]{fontenc}    
\usepackage{hyperref}       
\usepackage{url}            
\usepackage{booktabs}       
\usepackage{amsfonts}       
\usepackage{amsmath}
\usepackage{nicefrac}       
\usepackage{microtype}      
\usepackage{xcolor}         

\title{Augmenting Ground-Level PM2.5 Prediction via Kriging-Based Pseudo-Label Generation}

\author{%
  Lei Duan\thanks{Equal contribution} \\
  Department of Civil and \\ Environmental Engineering\\
  Duke University\\
  Durham, NC 27705 \\
  \texttt{lei.duan@duke.edu} \\
  \And
  Ziyang Jiang$^{*}$  \\
  Department of Civil and \\ Environmental Engineering \\
  Duke University \\
  Durham, NC 27705 \\
  \texttt{ziyang.jiang@duke.edu} \\
  \And
  David Carlson \\
  Department of Civil and \\ Environmental Engineering \\
  Duke University \\
  Durham, NC 27705 \\
  \texttt{david.carlson@duke.edu} \\
}

\begin{document}

\maketitle

\begin{abstract}
 Fusing abundant satellite data with sparse ground measurements constitutes a major challenge in climate modeling. To address this, we propose a strategy to augment the training dataset by introducing unlabeled satellite images paired with pseudo-labels generated through a spatial interpolation technique known as ordinary kriging, thereby making full use of the available satellite data resources. We show that the proposed data augmentation strategy helps enhance the performance of the state-of-the-art convolutional neural network-random forest (CNN-RF) model by a reasonable amount, resulting in a noteworthy improvement in spatial correlation and a reduction in prediction error.
\end{abstract}

\section{Introduction}
\label{intro}

Long-term exposure to fine particulate matter with an aerodynamic diameter of 2.5 $\mu m$ or smaller (PM$_{2.5}$) is is extensively recognized for increasing the risk of a variety of health problems such as stroke \cite{alexeeff2021long,yuan2019long}, respiratory diseases \cite{pope_health_2006}, and ischemic heart disease \cite{alexeeff2021long,thurston2016ischemic}. Accurate estimation and prediction of ground-level PM$_{2.5}$ concentrations are critical initial steps for subsequent epidemiological investigations into its health impacts. In recent years, advancements in satellite sensing techniques have enabled researchers to construct a high-resolution spatial mapping of PM$_{2.5}$ using machine learning models, in conjunction with ground measurements (e.g. PM$_{2.5}$, PM$_{10}$, temperature, humidity, etc.) acquired from air quality monitoring (AQM) stations \cite{zheng2020estimating,zheng_local_2021,jiang2022incorporating}. However, in most cases, the satellite-based data is much more abundant compared to ground measurements, resulting in a large amount of unlabeled satellite data (i.e., satellite images at locations without a paired ground measurement). To elaborate, satellites have the capacity to cover entire urban areas and a majority of suburban regions of a city within a 24-hour window. In contrast, measurements from ground stations are only representative within their immediate vicinity, which we refer to as the ``area of interest" (AOI). In most cases, the combined AOI of all ground stations is sparse compared to area covered by the satellite. 

To address the aforementioned limitation, we come up with an approach to make full use of the unlabeled satellite data and effectively integrate them with ground measurements. Specifically, we propose to pair the unlabeled satellite data with \emph{pseudo-labels} of ground measurements generated by a spatial interpolation method known as \emph{ordinary kriging} and then augment the labeled dataset using these paired instances. In this study, we adopt the convolutional neural network-random forest (CNN-RF) model from \citet{zheng_local_2021}. Our experimental results demonstrate the substantial advantages gained from this kriging-based augmentation, including decent reduction in errors and improvement in spatial correlation.

\section{Related work}
\label{work}

\paragraph{Kriging}
The theoretical basis of kriging, alternatively known as Gaussian process (GP) regression, was developed by Georges Matheron \cite{matheron1960krigeage}. Within geostatistics models, kriging serves as a spatial interpolation method where each data point is treated as a sample from a stochastic process \cite{oliver_kriging_1990,banerjee2003hierarchical,stein1999interpolation}. It also includes several distinct variants (e.g., ordinary kriging \cite{cressie1988spatial,wackernagel2003ordinary}, universal kriging \cite{stein1991universal}, co-kriging \cite{stein1991universal,helterbrand1994universal}, etc.) that are categorized based on the properties of the stochastic process. There are previous research efforts where kriging is applied to the context of remote sensing and climate modeling. To elaborate, \citet{zhan2018satellite} developed a random forest-spatiotemporal kriging model to estimate the daily NO$_{2}$ concentrations in China from satellite data and geographical covariates. \citet{wu2013spatial} employed residual kriging to interpolate average monthly temperature using latitude, longitude, and elevation as input variables. Besides these, researchers also used GPs for interpolation along with other sequence models \cite{futoma2017learning} or classification models \cite{li2016scalable} in machine learning.

\paragraph{Fusing Satellite and Ground Data}
The large gap in data density between satellite observations and ground measurements is a common phenomenon in the fields of remote sensing and climate modeling, leading researchers to develop a variety of approaches to address this issue. For instance, \citet{jiang2022improving} introduced the spatiotemporal contrastive learning strategy, which involves pre-training deep learning models using unlabeled satellite images to improve the model performance on ground measurements. \citet{verdin2015ab} employed Bayesian kriging to model the ground observations of rainfall, with the mean parameterized as a linear function of satellite estimates and elevation. In light of these prior studies, our proposed method shares similarities with both of these approaches.

\section{Methods}
\label{methods}

Our proposed method can be divided into 3 steps. First, we generate pseudo-labels of ground measurements using ordinary kriging and pair them with unlabeled satellite images. Then we integrate these satellite images with an existing training dataset containing true ground measurements. Lastly, we adopt the CNN-RF model from \citet{zheng_local_2021} to train on this augmented training dataset and compare the performance to the case without introducing pseudo-labels. We will discuss these steps in detail in the following subsections.


\begin{figure}[t!]
  \centering
  \includegraphics[width=\textwidth]{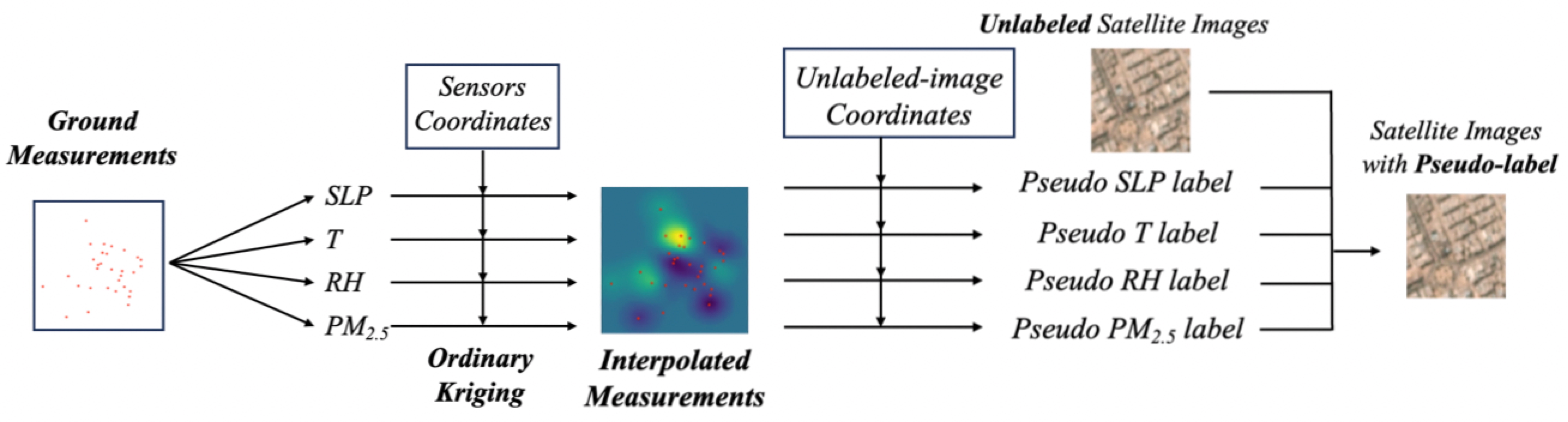}
  \caption{Illustration of pseudo-label generation process. The ground measurements consist of the ground-level PM$_{2.5}$ along with 3 meteorological attributes: sea level pressure (SLP), temperature (T), and relative humidity (RH). We first create a spatial mapping of interpolated measurements for the entire area of study using \emph{ordinary kriging}. Then we pair unlabeled satellite images with corresponding interpolated measurements based on their geographical coordinates. Namely, each final data point with pseudo labels consists of 5 components: satellite image, interpolated SLP, T, RH, and PM$_{2.5}$ measurements.}
  \label{fig:1}
\end{figure}


\subsection{Pseudo-label Generation}
\label{sec:3.1}
The process of generating pseudo-labels is illustrated in Figure \ref{fig:1}, where we use ordinary kriging with an exponential semivariogram (analogous to the kernel function in GP) to interpolate the ground measurements over the entire area of study. We prefer kriging over other spatial interpolation methods because it serves as an \emph{optimal linear predictor} under the modeling framework, accounting for factors such as spatial distance, continuity, and data redundancy. A more detailed discussion will be provided in Appendix \ref{appx:A1}. The parametric form for the exponential semivariogram is presented as follows:
\begin{equation}
\gamma(h) = 
\begin{cases}
\tau^2 + \sigma^2 (1 - \exp (-\phi h)) \quad &\text{if} \quad h > 0, \\
0 \quad &\text{otherwise},
\end{cases}
\end{equation}
where $h$ stands for the distance between two ground measurements, while $\tau^2$, $\tau^2 + \sigma^2$, and $1/\phi$ represent the nugget, sill, and range parameters, respectively. Concepts and details of semivariogram will be discussed in Appendix \ref{appx:A2}. We choose this particular form due to its superior fit with the empirical semivariogram compared to other isotropic semivariograms (e.g., Gaussian, spherical, etc.), as demonstrated in Appendix \ref{appx:A3}. Following the generation of pseudo-labels, we pair them with unlabeled satellite images based on their geographical coordinates. We then blend these pseudo-labeled images with existing satellite images with true ground measurements to form an \emph{augmented} training dataset. 
It is also possible to include uncertainty during imputation, but we only use the mean here as prior work shows that high-quality imputations make the biggest impact on performance \cite{li2016scalable}.

\subsection{Training of CNN-RF Joint Model}
With the augmented dataset, we proceed to the phase of model training. We adopt the convolutional neural network-random forest (CNN-RF) model from a previous study conducted by \citet{zheng_local_2021} as it demonstrates the state-of-the-art performance on the PM$_{2.5}$ prediction task. Specifically, the RF learns a preliminary estimation of PM$_{2.5}$ levels $\hat{y}_{\text{RF}}$ from the meterological attributes (i.e., SLP, T, RH) and the CNN learns a residual $\epsilon = y - \hat{y}_{\text{RF}}$ between the true PM$_{2.5}$ label $y$ and the RF-predicted PM$_{2.5} $ $\hat{y}_{\text{RF}}$. The model training phase is divided into two key components: Random Forest (RF) and Convolutional Neural Network (CNN). The detailed model architecture and hyperparameters can be found in Figure 2 in \citet{zheng_local_2021}.


\subsection{Model evaluation}
To ensure a direct comparative analysis, we test the effectiveness of our method using the same evaluation metrics adopted by \citet{zheng_local_2021}, which include the root-mean-square error (RMSE), mean absolute error (MAE), Pearson correlation coefficient (Pearson R), and spatial Pearson correlation coefficient (spatial Pearson R) as defined as follows:
\begin{align}
\text{RMSE} &= \textstyle \sqrt{\frac{1}{N}\sum_{i=1}^{N}(y - \hat{y})^2}, &
\text{MAE} &= \textstyle \frac{1}{N} \sum_{i=1}^{N} |y - \hat{y}| \\ 
\text{Pearson R} &= \textstyle \frac{\sum{(y - \bar{y})(\hat{y} - \bar{\hat{y}})}}{\sqrt{\sum{(y - \bar{y})^2} \sum{(\hat{y} - \bar{\hat{y}})^2}}},
&
\text{Spatial Pearson R} &= \textstyle  \frac{\sum{(y^{'} - \bar{y^{'}})(\hat{y} - \bar{\hat{y}})}}{\sqrt{\sum{(y - \bar{y})^2} \sum{(\hat{y} - \bar{\hat{y}})^2}}}
\end{align}
where $y$, $\hat{y}$ are the true PM$_{2.5}$ values and predicted PM$_{2.5}$ values, and $\bar{y}$,$\bar{\hat{y}}$ are the average true $PM_{2.5}$ values and the average predicted PM$_{2.5}$ values, $N$ is number of data points in the test set.

\begin{figure}[t!]
\begin{minipage}[c]{0.68\textwidth}
\includegraphics[width=\textwidth]{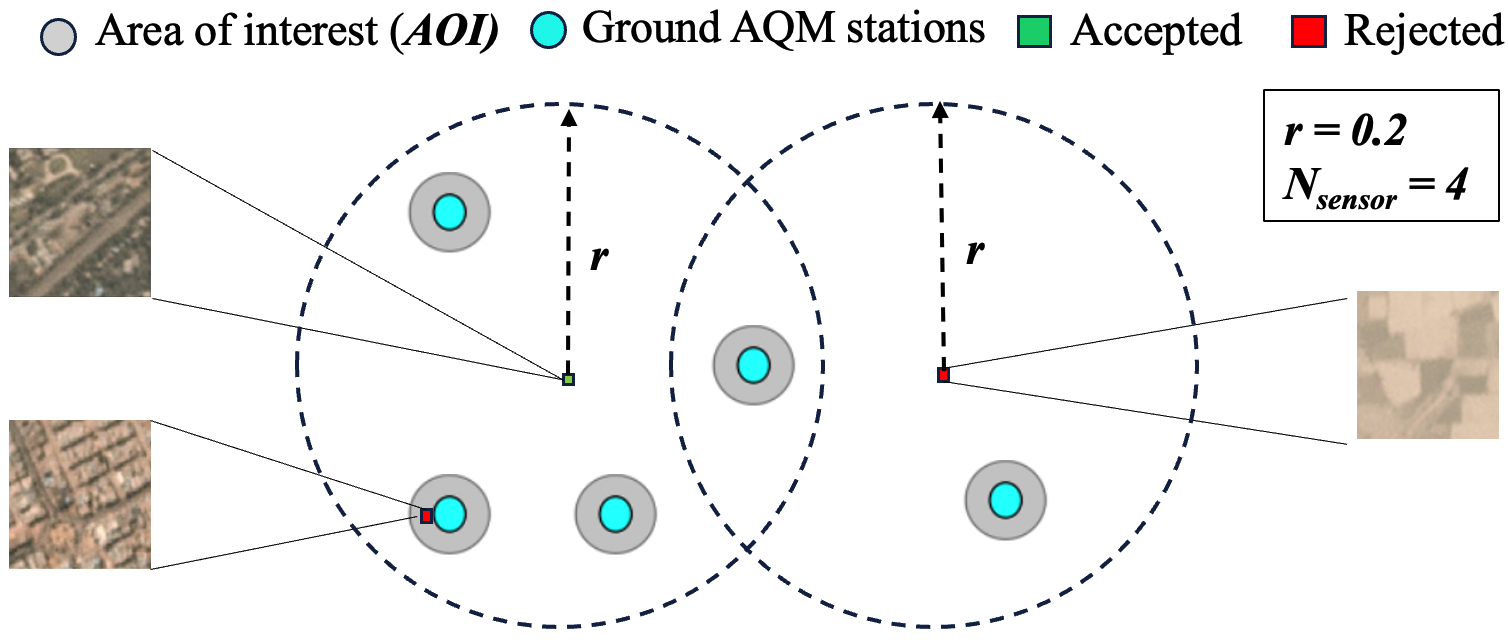}
\end{minipage}\hfill
\begin{minipage}[c]{0.29\textwidth}
\caption{Filtering rules for unlabeled satellite images. To be included in the training dataset, an unlabeled image must not be located within the AOI of any AQM station and must have at least $N_{\text{sensor}}$ AQM stations within its vicinity (defined by a circular region with a radius of $r$). In our experiments, we set $N_{\text{sensor}} = 4$ and $r = 0.2$.
} 
\label{fig:2}
\end{minipage}
\vspace{-3mm}
\end{figure}

\section{Experiments}
\label{result}

\subsection{Data}
\label{sec:4.1}
 We download the three-band (red-blue-green, RGB) scene visual products developed by Planet Labs \cite{planet} with a spatial resolution of 3 m/pixel, covering the period from January 1, 2018, to June 28, 2020. The existing training dataset contains 31,568 satellite images with true ground-level PM$_{2.5}$ labels and meteorological attributes obtained from 51 ground AQM stations spanning the National Capital Territory (NCT) of Delhi and its satellite cities such as Gurgaon, Faridabad, Noida, etc. We filter the unlabeled satellite images (i.e., images outside the AOI of AQM stations) as illustrated in Figure \ref{fig:2}. This procedure ensures that each image can be effectively paired with pseudo-labels that are spatially interpolated using a sufficient amount of surrounding ground measurement data. 
 

\subsection{Results}
\begin{figure}[t!]
  \centering
  \includegraphics[width=\textwidth]{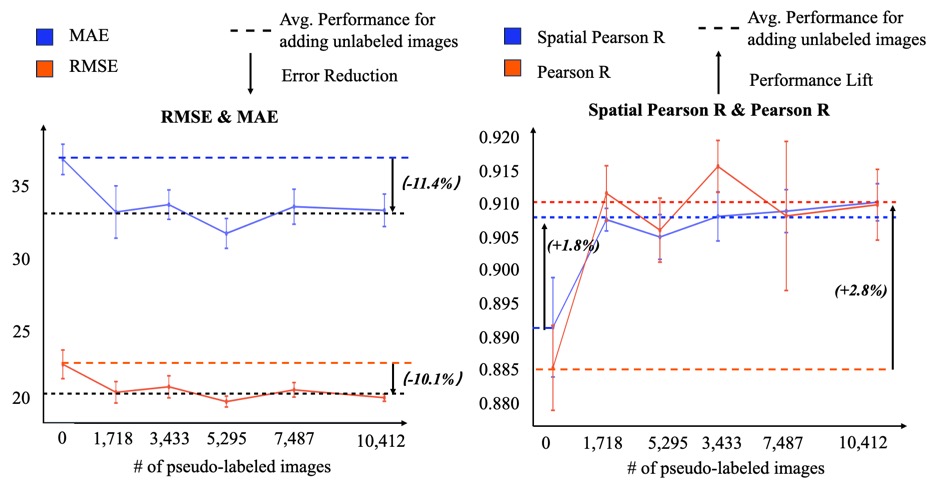}
  \caption{Performance of CNN-RF model on the test data with different number of pseudo-labeled images. (a) Root Mean Square Error (RMSE) and Mean Absolute Error (MAE), (b) Pearson R and Spatial Pearson R.}
  \label{fig:3}
\end{figure}

We test the CNN-RF model on datasets containing different number of pseudo-labeled satellite images and compare them with the baseline case with only true-labeled images (i.e., case with 0 pseudo-labeled images). For each scenario, we repeat the experiment by 10 times and calculate the average and standrd deviation of the evaluation metrics. As illustrated in Figure \ref{fig:3}, our results show substantial improvements in both prediction accuracy and spatial correlation. Specifically, we observe a 10.1\% reduction in average root mean square error (RMSE), decreasing from 37.64 to 32.56, and a 11.4\% decrease in mean absolute error (MAE), dropping from 24.39 to 20.31. Moreover, there is a 2.8\% increase in Pearson's correlation coefficient (Pearson R) from 0.880 to 0.912, and a 1.8\% increase in spatial Pearson's correlation coefficient (spatial Pearson R) from 0.878 to 0.907. These results highlight the substantial benefits gained from the integration of pseudo-labeled images within the training process.

\section{Conclusion}
In this research, we introduce a data augmentation approach that involves the incorporation of pseudo-labeled images into the model training procedure, allowing us to fully leverage the abundancy of unlabeled satellite images. Our approach involves the generation of pseudo-labeled images through a two-step process: firstly, applying ordinary kriging on ground measurements to produce pseudo-labels, and subsequently, pairing these labels with unlabeled images that are carefully selected through a heuristic criterion, ensuring that each unlabeled image has sufficient spatial information for spatial interpolation. The results demonstrate that including pseudo-labeled images successfully improve the state-of-the-art model performance in terms of prediction error and spatial correlation on the test data, showing the effectiveness of our proposed data augmentation strategy. We expect this strategy to be useful for various other scenarios in the field of remote sensing and geostatistics.

\bibliographystyle{unsrtnat}
\bibliography{references.bib}

\newpage
\appendix

\section{Details of Kriging}
\renewcommand\thefigure{\thesection\arabic{figure}}
\setcounter{figure}{0}
\label{appx:A}
\subsection{Advantages of Kriging}
\label{appx:A1}
There are several spatial interpolation methods in the existing geostatistics literature, such as bilinear interpolation, nearest neighbor, inverse distance weighting (IDW), modified Shepard interpolation etc. However, all these interpolation methods have certain limitations. Specifically, nearest neighbor approach gives very poor estimation when the ground stations are sparse. Both IDW and modified Shepard compute the estimated measurement at a target location as a weighted linear combination of its neighboring stations, with the weights being a nonlinear function of the distance $h$ between the target location and the neighboring station. A closer station (i.e., with smaller $h$) typically receives larger weights. Two commonly used weight functions are the inverse power function $w(h) = 1/h^{\beta}$ and the exponential function $w(h) = \exp \left(-(h/h_0)^{\beta}\right)$ where $\beta$ and $h_0$ are pre-determined parameters. 

However, a major drawback of IDW and modified Shepard lies in their failure to account for data redundancy. For instance, consider a scenario where we have measurements from ground stations denoted as $y_1, y_2, ..., y_m, y_{m+1}$, with the measurement we seek to estimate labeled as $y_0$. Let's assume that $y_1, y_2, ..., y_m$ are closely clustered together, while $y_{m+1}$ stands alone and further assume all of the stations, namely $y_1, y_2, ..., y_m, y_{m+1}$, are equidistant from $y_0$. With IDW or modified Shepard, $\{y_1, ..., y_m\}$ will receive much larger weights \emph{as a cluster} compared to $y_{m+1}$. Nevertheless, in reality, the cluster ${y_1, ..., y_m}$ contributes roughly the same amount of information to the estimation of $y_0$ as $y_{m+1}$ does. Therefore, it is reasonable to expect that $\{y_1, ..., y_m\}$ should receive similar weighting as $y_{m+1}$.

Kriging is a linear estimation method that minimizes the expected squared error, and therefore is also referred to as the \emph{best linear unbiased estimator}. To be specific, say we again have measurements from ground stations denoted as $y_1, y_2, ..., y_m$, with the measurement we want to estimate labeled as $y_0$. With Kriging, our estimation is formulated as a linear combination, denoted as $\hat{y}_0 = \sum_{i=1}^{m} \alpha_i y_i$, subject to the constraint $\sum_{i=1}^{m} \alpha_i = 1$. The determination of the coefficients $\alpha_i$ is closely tied to a function known as the \emph{variogram}, which will be elaborated in the following section. Compared to the aforementioned methods, Kriging takes into account spatial distance, spatial continuity, and data redundancy simultaneously. Consequently, it is the preferred method for our research, as it offers a more comprehensive approach to spatial estimation.

\subsection{Variogram} 
\label{appx:A2}
The variogram can be viewed as a function that describes the correlation between two ground measurements as a function of the distance $h$ between them. Mathematically, the semivariogram, which is half the value of the variogram, can be expressed by the general formulation as given below:
\begin{equation}
\gamma(h) = \frac{1}{2} \, \mathbb{E}[y(s+h) - y(s)]^2,
\end{equation}
where $y(s)$ and $y(s+h)$ represent the ground measurements at location $s$ and $s+h$, respectively. Some candidates of variogram include spherical, Gaussian, and exponential, whose functional forms are given as:

\begin{align}
\gamma_{\text{spherical}}(h) &= 
\begin{cases}
\tau^2 + \sigma^2 \quad &\text{if} \quad h \geq 1/\phi \\
\tau^2 + \sigma^2 \left[ \frac{3\phi h}{2} - \frac{1}{2} (\phi h)^{3} \right] \quad &\text{if} \quad 0 < h \leq 1/\phi \\
0 \quad &\text{otherwise}
\end{cases} \\
\gamma_{\text{Gaussian}}(h) &= 
\begin{cases}
\tau^2 + \sigma^2 \left( 1 - \exp (-\phi^2 h^2) \right) \quad &\text{if} \quad h > 0 \\
0 \quad &\text{otherwise}
\end{cases} \\
\gamma_{\text{exponential}}(h) &= 
\begin{cases}
\tau^2 + \sigma^2 (1 - \exp (-\phi h)) \quad &\text{if} \quad h > 0, \\
0 \quad &\text{otherwise},
\end{cases}
\end{align}

where $\tau^2$, $\tau^2 + \sigma^2$, and $1/\phi$ represent the nugget, sill, and range parameters, respectively. Here the nugget $\tau^2$ represents the semivariogram value at an infinitesimally small separation distance, i.e., $\tau^2 = \gamma(0^{+}) \equiv \lim_{h \rightarrow 0^{+}} \gamma(h)$. The range $1/\phi$ is the distance at which $\gamma(h)$ first reaches its ultimate level, i.e., the sill $\tau^2 + \sigma^2$.

\subsection{Empirical semivariogram}
\label{appx:A3}
\begin{figure}[t!]
\centering
\includegraphics[width=\textwidth]{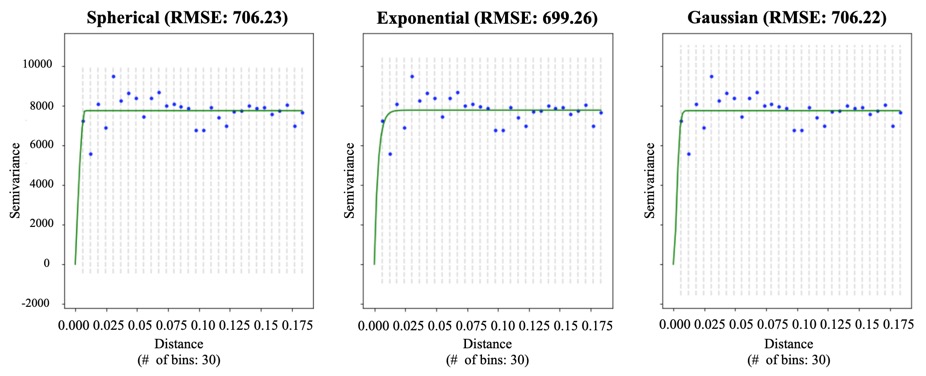}
\caption{Empirical semivariograms (plotted as blue dots) with different theoretical fits (plotted as green lines), i.e., spherical, exponential, and Gaussian.}
\label{fig:A1}
\end{figure}

As we elaborated in Section \ref{sec:3.1}, we plot the \emph{empirical semivariogram}:
\begin{equation}
\hat{\gamma}(h) = \frac{1}{2N(h)} \sum_{(s_i, s_j) \in N(h_k)} \left[ y(s_i) - y(s_j) \right]^2, 
\end{equation}
where $N(h_k)$ is the set of pairs of points $(s_i, s_j)$ such that $\| s_i - s_j \| \in I_k$. Here $I_k$ represents the interval $(h_{k-1}, h_k)$ such that $0 < h_1 < h_2 < \cdots < h_K$ and $k \in \{1, 2, ..., K\}$. Figure \ref{fig:A1} provides a visual comparison between the empirical semivariograms computed from ground measurements and the theoretical forms of spherical, exponential, and Gaussian semivariograms. Among these models, the exponential semivariogram exhibits the lowest RMSE when compared to the other two models. The RMSE between the empirical semivariogram and the theoretical model is calculated as:
\begin{equation}
\text{RMSE} = \sqrt{\frac{1}{N_{\text{bins}}} \sum_{i=0}^{N_{\text{bins}}} \left( \hat{\gamma}_i - \gamma_i \right)^2}, 
\end{equation}
where $N_{\text{bins}}$ is the total number of bins used for plotting the empirical semivariogram, $\hat{\gamma}_i$ is the empirical semivariogram value for the $i^{th}$ bin, and $\gamma_i$ is the theoretical value for the $i^{th}$ bin.

\end{document}